\documentclass[lettersize,journal,twoside]{IEEEtran} % 'twoside' is required for correct page headers
\usepackage{amsmath,amsfonts,amssymb}
\usepackage{algorithmic}
\usepackage{algorithm}
\usepackage{array}
\usepackage{enumitem}
\usepackage{textcomp}
\usepackage{stfloats}
\usepackage{url}
\usepackage{verbatim}
\usepackage{graphicx}
\usepackage{bm}
\usepackage{cite}
\usepackage{booktabs}
\usepackage{tablefootnote}
\usepackage{multirow}
\usepackage{colortbl}
\usepackage{diagbox}
\usepackage{threeparttable}
\usepackage{mathtools}
\usepackage{subfigure}
\usepackage[dvipsnames]{xcolor}
\usepackage{color}
\definecolor{linkblue}{rgb}{0, 0.19, 0.32}
\usepackage[colorlinks, linkcolor=black]{hyperref}
\hypersetup{
    colorlinks=true, %设置链接的颜色
    urlcolor=linkblue
    }

\definecolor{myblue}{RGB}{0, 0, 255}

\begin{document}

\title{Analyzing Key Objectives in Human-to-Robot Retargeting for Dexterous Manipulation}

\author{
Chendong Xin\textsuperscript{*},
        Mingrui Yu\textsuperscript{*},
        Yongpeng Jiang,
        Zhefeng Zhang,
        and Xiang Li\textsuperscript{$\dagger$}
        % IEEE Publication Technology,~\IEEEmembership{Staff,~IEEE,}
        % <-this % stops a space
\thanks{
\textsuperscript{*}Equal contribution.}%
\thanks{
\textsuperscript{$\dagger$}Corresponding author: \texttt{xiangli@tsinghua.edu.cn}.}%
\thanks{C. Xin, M. Yu, Y. Jiang, Z. Zhang, and X. Li are with the Department of Automation, Beijing National Research Center for Information Science and Technology, Tsinghua University, China.
This work was supported in part by Science and Technology Innovation 2030-Key Project under Grant 2021ZD0201404, in part by the National Natural Science Foundation of China under Grant 62461160307 and 623B2059, in part by the Fundamental and Interdisciplinary Disciplines Breakthrough Plan of the Ministry of Education of China under Grant JYB2025XDXM208, in part by the BNRist project under Grant BNR2024TD03003, and in part by the Institute for Guo Qiang, Tsinghua University.
}%
}

% The paper headers
% \markboth{Journal of \LaTeX\ Class Files}%
% {Shell \MakeLowercase{\textit{et al.}}: A Sample Article Using IEEEtran.cls for IEEE Journals}

% \IEEEpubid{0000--0000/00\$00.00~\copyright~2021 IEEE}
% Remember, if you use this you must call \IEEEpubidadjcol in the second
% column for its text to clear the IEEEpubid mark.

\maketitle

\begin{abstract}
Kinematic retargeting from human hands to robot hands is essential for transferring dexterity from humans to robots in manipulation teleoperation and imitation learning. 
However, due to mechanical differences between human and robot hands, completely reproducing human motions on robot hands is impossible. 
Existing works on retargeting incorporate various optimization objectives, focusing on different aspects of hand configuration. However, the lack of experimental comparative studies leaves the significance and effectiveness of these objectives unclear.
This work aims to analyze these retargeting objectives for dexterous manipulation through extensive real-world comparative experiments. 
Specifically, we propose a comprehensive retargeting objective formulation that integrates intuitively crucial factors appearing in recent approaches. 
The significance of each factor is evaluated through experimental ablation studies on the full objective in kinematic posture retargeting and real-world teleoperated manipulation tasks. 
Experimental results and conclusions provide valuable insights for designing more accurate and effective retargeting algorithms for real-world dexterous manipulation. Supplementary materials are available at \href{https://mingrui-yu.github.io/retargeting}{https://mingrui-yu.github.io/retargeting}.
% To address this challenge, existing retargeting approaches designed algorithms with task-oriented objectives to adapt to different manipulation tasks. 
% The design of retargeting algorithms should focus on the demand of certain manipulation tasks and include relative objectives for better manipulation performance. 
% Existing retargeting approaches vary in their choices of objectives to adapt to different applications. 
% This work summarizes retargeting objectives that are commonly used in recent researches and proposes a complete retargeting objective formulation with a series of key factors. 
% Specifically, we propose a complete retargeting objective formulation with a series of key factors commonly used in recent research.
% The significance of these factors on different manipulation tasks are analyzed and concluded through extensive experimental studies on kinematic posture retargeting and real-world manipulation tasks. 
\end{abstract}


\section{Introduction}

\IEEEPARstart{T}{ransferring} dexterity from humans to robots is a promising field in dexterous manipulation research, as the complexity of dexterous manipulation tasks poses challenges for classical analytical approaches \cite{pang2023global,jiang2024contact,yu2025robotic,yu2024hand}. 
In either collecting robot demonstrations through human teleoperation or learning from offline human manipulations, an essential component is \textit{kinematic retargeting}, which refers to kinematically translating the human hand configuration to robot hand joint positions.
One key challenge of retargeting is that, due to the differences in morphology and degrees of freedom (DoFs) between human hands and current robot hands, the human hand configurations can not be exactly reproduced by the robot hand.
Consequently, it is inevitable to focus on only partial configuration of the human hand that are more crucial to manipulation tasks when designing the retargeting algorithms.

% Dexterous retargeting is challenging due to significant differences in morphology and degrees of freedom (DoF) between human and robot hands. 
% Moreover, dexterous manipulation requires a high level of precision and adds to the complexity of retargeting.

% \IEEEPARstart{T}{eleoperation} systems play a crucial role in human-robot interaction, allowing human operators to remotely control robots in real time. 
% % These systems have been widely applied in tasks such as telesurgery, remote maintenance, and industrial assembly. 
% With the increasing integration of learning-based approaches in robotics, teleoperation has been used extensively to collect demonstration data to facilitate imitation learning. 
% Dexterous teleoperation, in which a multi-fingered dexterous robot hand commonly serves as the end-effector of a robot arm, is specifically used to execute and collect demonstrations for dexterous manipulation tasks. 

% In dexterous teleoperation, a key component is kinematic retargeting, which translates human hand poses to robot hand. However, this process is challenging due to significant differences in morphology and degrees of freedom (DoF) between human and robot hands. Additionally, dexterous manipulation requires a high level of precision and adds to the complexity of retargeting.

% Many dexterous teleoperation systems employ retargeting methods with various objectives to minimize the discrepancy between human and robot hand movements. Based on the representation of human hands, these methods can be broadly categorized into joint-space and Cartesian-space approaches.

\begin{figure}
    \centering
    \includegraphics[width=0.95\linewidth]{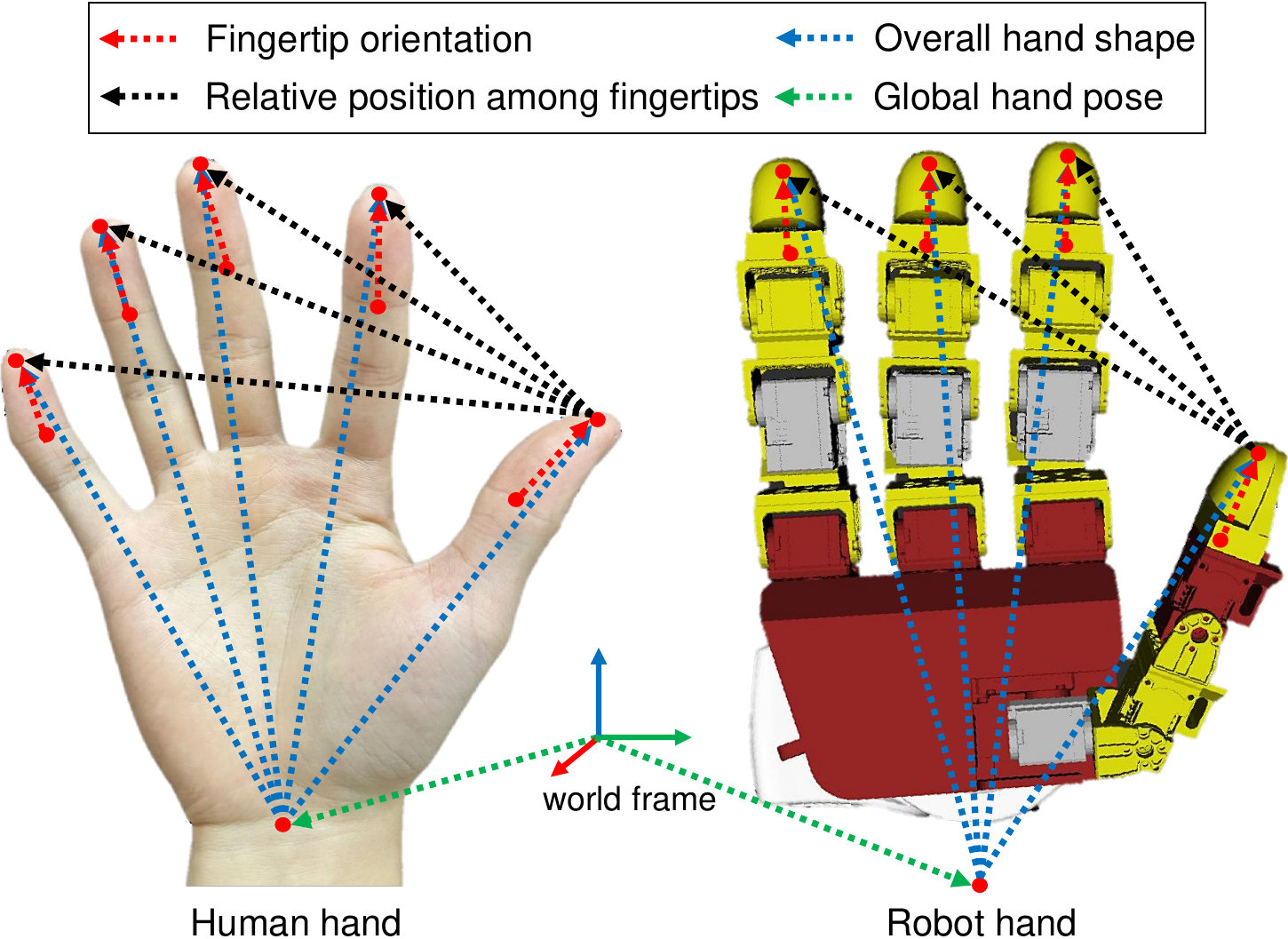}
    \vspace{-2mm}
    \caption{
     Crucial objectives in human-to-robot retargeting for dexterous manipulation. This work explores the appropriate formulation of these objectives and experimentally analyzes their significance for different manipulation tasks.
    }
    \label{fig:hand}
    \vspace{-2mm}
\end{figure}

The most straightforward retargeting approach is direct joint-to-joint mapping, where the robot hand joints follow a manually defined transformation of the human hand joints \cite{gao2023hand, huang2025dih, wei2023adaptive, guo2024telephantom, liarokapis2013telemanipulation, giudici2023feeling, arunachalam2023holo, iyer2024open}. These approaches require considerable manual efforts to define the joint-space transformation for a certain robot hand.
Task-space retargeting is used more widely, which relies on inverse kinematics (IK) solving to determine robot hand joint configuration based on human hand keypoints. 
The global fingertip position is the fundamental retargeting objective to align human and robot hand in task-space \cite{wang2024dexcap, shaw2024bimanual, zhang2025doglove, arunachalam2023dexterous, chen2024arcap}.
% In most works, the relative fingertip position to the wrist serves as the retargeting objective \cite{antotsiou2018task, qin2022dexmv, qin2023anyteleop, ding2024bunny, yang2024ace, huang2024designing, shaw2024learning, yuan2024cross, cheng2024open, romero2024eyesight, handa2020dexpilot, sivakumar2022robotic}. Distance between two set of vectors from the wrist to the fingertips is minimized to obtain the robot’s optimal joint configurations, thereby preserving the overall hand shape. 
Most works use vectors from wrist to fingertips as the retargeting objective and minimize their discrepancy between human and robot hands to preserve overall hand shape\cite{antotsiou2018task, qin2022dexmv, qin2023anyteleop, ding2024bunny, yang2024ace, huang2024designing, shaw2024learning, yuan2024cross, cheng2024open, romero2024eyesight, handa2020dexpilot, sivakumar2022robotic}.
% Some works include finger orientations in the objective to preserve more human-like shapes. The finger shapes are represented in the form of finger bending information by including a set of vectors from palm to middle phalanx \cite{qin2022dexmv}. More explicitly, the directions of five proximal phalanges and distal phalanges \cite{enwiki:1275529475} are used to represent orientations of the finger-roots and fingertips \cite{li2019vision}. 
Some works further incorporate fingertip orientation into the objective, using vectors from the palm to the middle phalanx to represent finger bending information\cite{qin2022dexmv}, or using the directions of proximal phalanges and distal phalanges \cite{enwiki:1275529475} to represent orientations of the finger-roots and fingertips \cite{li2019vision}.
Additionally, relative position among fingertips can be included to address the challenge of pinch grasps \cite{handa2020dexpilot}, in which a switching weight is used to prioritize the fingers involved in pinching, and the corresponding reference distance on human hand is adjusted to a minimal distance to encourage fingertip contact.

\IEEEpubidadjcol

% Joint-space methods directly map human joint configurations to the robot hand, but due to differences in hand morphology and DoF, they often require manual adjustments to ensure feasibility and effectiveness.
% Cartesian-space methods, which are more commonly used, utilize human hand keypoint poses in Cartesian space to compute robot joint configurations. Basic position-based methods minimize the absolute position error between human and robot hand keypoints, while Task Space Vector (TSV)-based methods incorporate relative position information to preserve spatial relationships between hand segments.

These retargeting approaches have been deployed in real-world manipulation, but they differ in their choices of objectives. Due to the lack of comparative studies, it remains unclear which objectives are more dispensable for which types of manipulation tasks and whether conflicts exist between them.
A previous survey \cite{meattini2022human} reviews and classifies the common retargeting methods, but it does not focus on retargeting objectives and not provide experimental evaluations.

This work aims to analyze the key objectives in retargeting for dexterous manipulation through experimental comparison. 
We summarize objectives that are used in recent research on dexterous retargeting and propose a comprehensive retargeting objective formulation that considers all intuitively crucial factors illustrated in Fig. \ref{fig:hand}.
By conducting ablation studies on the full objective in kinematic posture retargeting and real-world manipulation teleoperation, we analyze the significance and effectiveness of different objectives in different tasks.
The experimental results and conclusions provide valuable insights for future research on designing retargeting algorithms for learning dexterous manipulation from humans or teleoperation.

\section{Method}

% We start from several factors that are intuitively crucial to retargeting and illustrate how to form them as components of a complete retargeting objective:
We intuitively list the factors that are potentially crucial to human-to-robot retargeting for dexterous manipulation:
\begin{itemize}
    \item \textbf{Global hand pose}: the robot's hand pose in the global frame should be aligned with the human's in dexterous manipulation involving large arm motions.
    \item \textbf{Overall hand shape}: the robot hand should replicate a similar overall hand shape to the human hand to achieve similar postures.
    \item \textbf{Relative position among fingertips}: the spatial relationship among each fingertip (e.g., thumb and index) is critical for manipulation tasks that require precise coordination of fingertips. 
    \item \textbf{Fingertip orientations}: the accurate fingertip orientations ensure appropriate directions of contact normals during contact-rich dexterous manipulation.
\end{itemize}

Based on the intuitive factors above, we formulate their corresponding mathematical representations and construct a complete retargeting objective.

\textbf{Global hand pose}: 
The global hand pose is typically formulated with a wrist position term $\mathcal{L}_{\text{wrist\_pos}}$ and a wrist orientation term $\mathcal{L}_{\text{wrist\_rot}}$. However, accurate tracking of global wrist pose is not necessary in most dexterous manipulation tasks and may reduce accuracy in fingertip positions.
Due to different hand morphologies between the human and robot, the fingertip positions relative to the wrist may conflict with the relative positions between fingertips and fingertip orientations. 
As a result, it can be better to allow adjustment of the wrist pose in exchange for higher fingertip accuracy.
% An accurate tracking of wrist pose may lead to an undesired higher sensitivity to minor arm movements which do not affect hand manipulation.
Specifically, we replace the wrist position objective with a thumb-tip position objective $\mathcal{L}_{\text{thumb\_pos}}$ and apply joint optimization of the arm-hand joint positions. 
% This allows for a better positioning of fingertips, which is more critical for dexterous manipulation. 
The retargeted wrist orientation is regularized through the wrist orientation objective $\mathcal{L}_{\text{wrist\_rot}}$ with a relatively small weight. The complete term is formulated as:
% while preserving an approximate control over the wrist pose through the wrist orientation term $\mathcal{L}_{\text{wrist\_rot}}$ and the wrist-to-tip vectors in other terms.
% \begin{equation}
%     \mathcal{L}_{\text{hand\_pose}} = \left\|p^{h}_{\text{thumb}}-p^{r}_{\text{thumb}}\right\|_2 + \beta_{\text{rot}}\cdot \arccos(2{\langle q^{h}_{\text{wrist}}, q^{r}_{\text{wrist}}\rangle}^{2}-1)
% \end{equation}
\begin{equation} \label{eq:global_pose}
    \mathcal{L}_{\text{hand\_pose}} = \left\|\bm{p}^{h}_{\text{thumb}}-\bm{p}^{r}_{\text{thumb}}\right\|_2^2 
    + \beta_{\text{rot}} \,
    \text{angle}( \bm q^{h}_{\text{wrist}}, \bm q^{r}_{\text{wrist}} ) ,
\end{equation}
% \vspace{-2mm}
where $\bm{p}_{\text{thumb}}$ is the thumb fingertip position of human and robot, and $\bm{q}_{\text{wrist}}$ is the orientation of human and robot wrist.

\textbf{Overall hand shape}: The overall hand shape is represented by the fingertip positions relative to the wrist position. The fingertip position term $\mathcal{L}_{\text{fingertip\_pos}}$ measures the difference in a set of vectors defined from the wrist to the fingertips of the robot and human hand:

\begin{equation}
     \mathcal{L}_{\text{fingertip\_pos}} = \sum_{i=1}^{N} \tilde{s}(d_{i})\left\| \bm{v}_i^{r} - \bm{v}_i^{h} \right\|^2,
\end{equation}

\noindent where $\bm{v}_i$ is the vector from the wrist to the $i^{th}$ fingertip on human and robot hand, and $N$ is the number of fingers. $\tilde{s}(d_{i})$ is a switching weight function to balance fingertip positions relative to the wrist and the thumb, where
\begin{equation*}
    \tilde{s}(d_{i}) = \operatorname{sigmoid}(d_{i}, \epsilon_{1}, -10), 
\end{equation*}
This term is designed to be coordinated with the following pinch objective $\mathcal{L}_{\text{pinch}}$, so that the sum of pinch term switching weight $s(d_{i})$ and the switching weight here $\tilde{s}(d_{i})$ will be a fixed number. 
Later, in ablation studies where the pinch term is removed (A1, A6 and A8), we set $\tilde{s}(d_{i})$ to be a constant $1.0$.

\textbf{Relative position among fingertips}: We use vectors from the thumb to primary fingers (index, middle, ring) to represent the relative positions among fingertips, which is crucial for pinching. 
% This consideration is based on the experience that the coordination between the thumb and the primary fingers is most crucial for precise manipulation tasks like pinching. 
We adopt a similar formulation to DexPilot \cite{handa2020dexpilot} with a switching weight function $s(d_{i})$ and a distance rescaling function $l(d_{i})$:

\begin{equation} \label{eq:pinch_term}
     \mathcal{L}_{\text{pinch}} = \sum_{i=1}^{N-1} s(d_{i})\left\| \bm{\gamma}_i^{r} - l(d_{i})\hat{\bm{\gamma}}_i^{h} \right\|_2^2 ,
\end{equation}
where $\bm{\gamma}_i$ is the vector from the thumb fingertip to the fingertip of the $i^{th}$ primary finger, $d_{i}=\left\|\bm{\gamma}_{i}^{h}\right\|$ and $\hat{\bm{\gamma}}_i^{h} = \frac{\bm{\gamma}_{i}^{h}}{d_{i}}$. Instead of using a discrete weight function as DexPilot, we use a continuous weight function to ensure smooth transitions:
\begin{equation*}
    s(d_{i}) = \operatorname{sigmoid}(d_{i}, \epsilon_{1}, 10),
\end{equation*}
where $\operatorname{sigmoid}(\cdot)$ is the sigmoid function defined as follows:
\begin{equation*}
    \operatorname{sigmoid(x, c, w)}=\frac{1}{1+e^{w(x-c)}}.
\end{equation*}
 
% The distance rescaling function $l(d_{i})=\frac{\epsilon_{1}}{\epsilon_{1} - \epsilon_{2}}(d_i - \epsilon_{2})$ linearly rescales human fingertip distances from $[\epsilon_{2},\epsilon_{1}]$ to $[0, \epsilon_{1}]$ to ensure a continuous transition in the pinching range and avoids sudden changes around the threshold $\epsilon_{1}$.
% Values below $\epsilon_{2}$ are clamped to zero 
% Detailed explanation and hyper-parameter choices are provided in Appendix.}
Our distance rescaling function is defined as follows:
\begin{equation}
l(d_{i}) = 
\begin{cases}
0, & d_i < \epsilon_{2} \\
\frac{\epsilon_{1}}{\epsilon_{1} - \epsilon_{2}}(d_i - \epsilon_{2}), & \epsilon_{2} \leq d_i \leq \epsilon_{1} \\
d_i, & d_i > \epsilon_{1} , 
\end{cases}
\end{equation}
where fingertip distance within pinching range $[\epsilon_{2},\epsilon_{1}]$ is linearly rescaled into $[0, \epsilon_{1}]$. This ensures a continuous transition in the pinching range and avoids sudden changes around the threshold $\epsilon_{1}$.
In practice we set $\epsilon_{1}=1\times10^{-1}\,\text{m}$ and $\epsilon_{2}=1\times10^{-2}\,\text{m}$. 

\textbf{Fingertip orientations}:
To represent fingertip orientations, we include another set of vectors.
% and formulate a fingertip orientation term $\mathcal{L}_{\text{fingertip\_rot}}$ . 
In contrast with methods such as DexMV \cite{qin2022dexmv} that define the vectors from the wrist to the middle phalanx, our formulation defines the vectors from the distal interphalangeal (DIP) joints to the fingertips, which represent fingertip orientations more directly:

\begin{equation} \label{eq:fingertip_rot_term}
    \mathcal{L}_{\text{fingertip\_rot}} = \sum_{i=1}^N \left\|\bm{r}_i^{h}-\bm{r}_i^{r}\right\|_2^2 ,
\end{equation}

\noindent where $\bm{r}_{i}$ is the vector from the DIP joint to the fingertip.
% on the $i^{th}$ finger.

The final retargeting optimization objective considering all the above factors can be specified as:
\begin{equation} \label{eq:full_objective}
\begin{aligned}
    \mathcal{L}_{\text{total}} =\; 
    & \lambda_1\mathcal{L}_{\text{thumb\_pos}} + 
    \lambda_2\mathcal{L}_{\text{wrist\_rot}} + 
    \lambda_3\mathcal{L}_{\text{fingertip\_pos}} + \\
    & \lambda_4\mathcal{L}_{\text{fingertip\_rot}} +
    \lambda_5\mathcal{L}_{\text{pinch}} + \mathcal{L}_{\text{joint}} + \mathcal{L}_{\text{vel}}, 
\end{aligned}
\end{equation}
where two joint-space regularization terms are additionally considered.
The joint position regularization $\mathcal{L}_{\text{joint}} = \sum_{j=1}^{m} w_j^{\rm pos} \left\| q_j - \bar{q}_j \right\|_2^2$ penalizes the deviation of some joints from a pre-defined joint configuration $\bar{\bm q}$. 
% to limit configurations of certain joints.
The joint velocity regularization $\mathcal{L}_{\text{vel}} = \sum_{j=1}^{m} w_{j}^{\rm vel} \left\| q_j - q_j^{\text{prev}} \right\|_2^2 $ penalizes large changes in joint positions compared to previous timestep to encourage trajectory smoothness. Its effect is illustrated by the joint position/velocity/acceleration profiles in Appendix.

% We conduct a series of ablation experiments based on the complete objective to analyze the contribution of each component term. We focus on the thumb fingertip position term $\mathcal{L}_{\text{thumb\_pos}}$, the fingertip orientation term $\mathcal{L}_{\text{fingertip\_rot}}$, the pinch term $\mathcal{L}_{\text{pinch}}$ and the joint position regularization term $\mathcal{L}_{\text{joint}}$. Ablations are performed by either removing certain term or modifying its formulation. This allows us to analyze the effect of each term independently.

\begin{table*}[t]
\centering
\caption{Definition of the ablations for comparative studies.}
\begin{tabular}{lll}
\toprule
\textbf{Category} & \textbf{Ablation} & \textbf{Definition} \\
\midrule
Full & \textcolor{Maroon}{Full} & Complete retargeting objective $\mathcal{L}_{\text{total}}$ with all terms (\ref{eq:full_objective}) \\
\midrule
\multirow{2}{*}{Fingertip pinch} 
     & \textcolor{blue}{A1} & Remove the pinch term $\mathcal{L}_{\text{pinch}}$ (\ref{eq:pinch_term}) \\
     & \textcolor{blue}{A2} & Use actual pinch distance without distance rescaling in (\ref{eq:pinch_term})\\
\midrule
\multirow{2}{*}{Fingertip orientation} 
     & \textcolor{ForestGreen}{A3} & Remove the fingertip orientation term $\mathcal{L}_{\text{fingertip\_rot}}$ (\ref{eq:fingertip_rot_term}) \\
     & \textcolor{ForestGreen}{A4} & Replace the vectors from DIP joints to fingertips with vectors from the wrist to DIP joints in (\ref{eq:fingertip_rot_term}) \\
\midrule
\multirow{2}{*}{Global wrist pose} 
     & \textcolor{violet}{A5} & Replace the thumb position term in (\ref{eq:global_pose}) with a wrist position term \\
     & \textcolor{violet}{A6} & Replace the thumb position term in (\ref{eq:global_pose}) with a wrist position term  and remove $\mathcal{L}_{\text{pinch}}$ and $\mathcal{L}_{\text{fingertip\_rot}}$ \\
\midrule
\multirow{2}{*}{Joint regularization}
     & \textcolor{orange}{A7} & Remove the joint position regularization $\mathcal{L}_{\text{joint}}$ \\
     &  \textcolor{orange}{A8} & Replace the thumb position term in (\ref{eq:global_pose}) with a wrist position term  and remove $\mathcal{L}_{\text{pinch}}$, $\mathcal{L}_{\text{fingertip\_rot}}$, and $\mathcal{L}_{\text{joint}}$ \\
\bottomrule
\end{tabular}
\label{tab:ablation}
\end{table*}

\section{Evaluations and Results}

\subsection{Evaluation Setup}

\textbf{Implementation}:
Simulation studies involve a Leap Hand \cite{shaw2023leaphand} and a Shadow Hand, both mounted on a Franka Emika Panda arm.
% we use a Franka Emika Panda robot arm equipped with a Leap Hand \cite{shaw2023leaphand} or a Shadow Hand.
% with five fingers and 20 actuated DoFs.
Real-world experiments are conducted on a teleoperation system with a Leap Hand, a Panda arm, and an Apple Vision Pro for human hand detection \cite{park2024avp}.
% For real-world experiments, we develop a teleoperation system on the Leap Hand and Panda arm. 
% For hand pose detection, we use the Vision Pro headset to detect human hand keypoints via \cite{park2024avp}.
The retargeting objective is optimized in real time using the Sequential Least-Squares Quadratic Programming in the NLopt library \cite{NLopt}.

In the real-world experiments, we use joint-space position control for both the Panda arm and the robot hand. At each retargeting step (20 Hz), the retargeting optimizer outputs a desired joint configuration $\bm{q}_t$ for all actuated joints. This configuration is first upsampled to a 100 Hz command stream via interpolation and then sent to the hardware. 
The LEAP Hand executes the control command via a built-in current-based PD controller.
The Franka arm runs an impedance controller at a higher 1000 Hz with strict acceleration and jerk limits, so the 100 Hz joint targets are further interpolated to match the 1000 Hz control frequency before being executed by the arm controller.

In Appendix, we further discuss the runtime and latency of our system, as well as its robustness to tracking noise and occlusions.

\textbf{Ablation setup}:
To analyze the significance of each retargeting objective, we implement ablations by removing or changing certain objective terms in $\mathcal{L}_{\text{total}}$ (\ref{eq:full_objective}).
% We evaluate the full retargeting objective with several ablations on a series of kinematic postures as well as real-world manipulation tasks. 
The detail of the ablation setup is described in Table \ref{tab:ablation}.

\begin{figure*}
    \centering
    \includegraphics[width=0.98\linewidth]{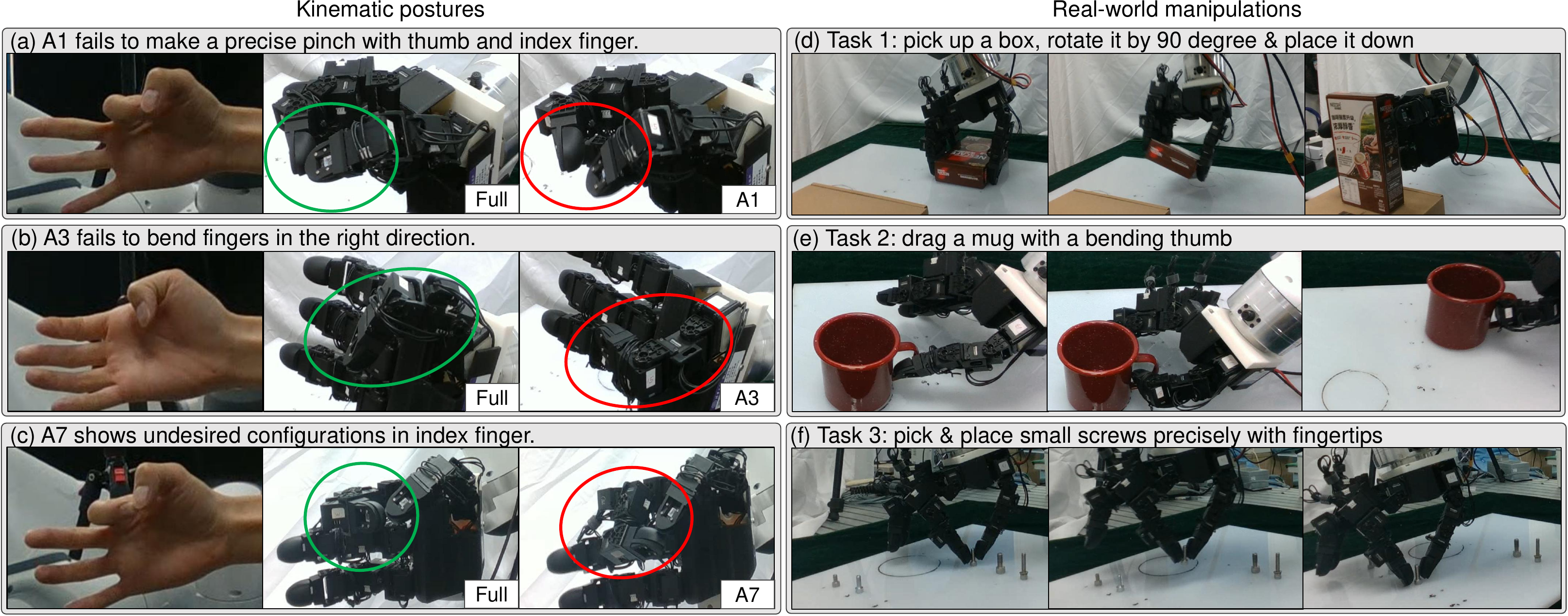}
    \vspace{-2mm}
    \caption{Snapshots of the real-world kinematic postures retargeting (left) and the three manipulation tasks (right). 
    (a) to (c): each row shows a human hand posture and the corresponding retargeted robot postures using the full objective and an ablation implementation.
    % Each figure of kinematic posture shows the same frame of human demonstration, full retargeting objective and an ablation respectively. Each of the ablation demonstrates certain deficiency in replicating posture of the human pilot. 
    (d) to (e): each row shows the snapshots of the manipulation process of one task using the full retargeting objective.}
    \label{fig:posture_and_task}
    % \vspace{-4mm}
\end{figure*}

\begin{figure*}[tb]
    \centering
    \includegraphics[width=\linewidth]{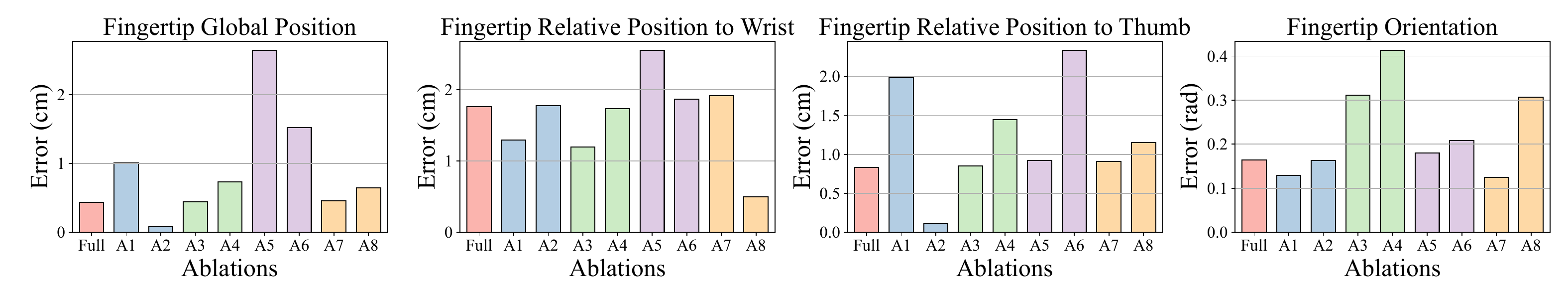}
    \vspace{-5mm}
    \caption{Results of kinematic posture retargeting on finger pinch trajectories. 
    Each bar shows the error of one ablation implementation and the colors represent the ablation category defined in Table \ref{tab:ablation}.
    % The leftmost bar shows the errors of the full objective, while the remaining bars show the errors of different ablations. 
    % Colors of the bars represent ablations in different categories defined in Table \ref{tab:ablation}.
    For the metrics of fingertip global position and fingertip relative position to the thumb, only the errors of the two fingers involved in the pinching motion are considered.}
    \label{fig:leap_simulation_1}
    % \vspace{-4mm}
\end{figure*}

\textbf{Kinematic posture retargeting tasks}:
We evaluate the full retargeting objective and eight ablations across offline data of three pinch motions involving the thumb and the index, middle, and ring fingers, respectively. 
We use four quantitative metrics for kinematic postures: 
1) average fingertip position error in the global frame, 
2) average fingertip position error relative to the wrist, 
3) average fingertip position error relative to the thumb, 
and 4) average fingertip orientation error. 
The results on Leap Hand are shown in Fig. \ref{fig:leap_simulation_1}. 
The results on another trajectory involving finger crossing motions and results on Shadow Hand are provided in the Appendix.
The snapshots of posture retargeting in the real world are shown in Fig. \ref{fig:posture_and_task}.

\begin{figure}[tb]
    \centering
    \includegraphics[width=\linewidth]{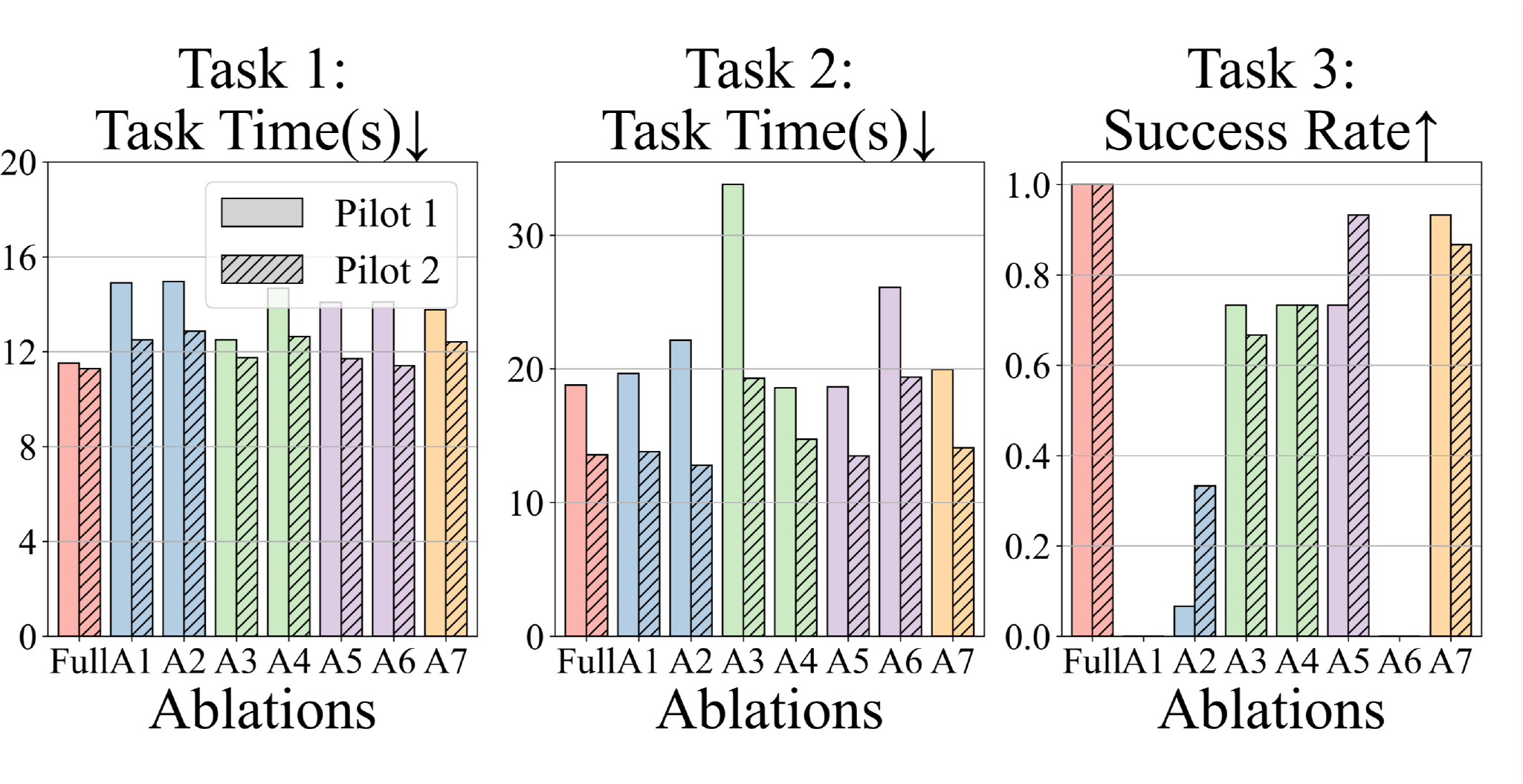}
    \vspace{-6mm}
    \caption{Results of the real-world manipulations. Task 1 and 2 are assessed by task time, while Task 3 is evaluated by success rate.
    Each pair of bars shows the error of one ablation implementation and the colors represent the ablation category defined in Table \ref{tab:ablation}.  The two pilots are distinguished by hatched bars.
    }
    \label{fig:real_world_results}
    % \vspace{-2mm}
\end{figure}

\textbf{Real-world manipulation tasks}:
We design three representative real-world manipulation teleoperation tasks : 
\begin{enumerate}[label=\arabic*), itemsep=0pt, topsep=0pt, parsep=0pt, partopsep=0pt]
    % \item Pick up a box, rotate it by 90 degrees, and place it down, which represents commonly used pick-and-place tasks in dexterous teleoperation.
    \item Pick up a box, rotate it, and place it down, representing common pick-and-place tasks in dexterous teleoperation.
    \item Drag a mug through its handle using the bent thumb, where fingertip orientation plays a decisive role.
    % \item Pick up five upright-standing screws of different sizes and place them on target positions vertically, where precise pinches are important.
    \item Pick up five different upright-standing screws and place them vertically, where precise pinches are important.
\end{enumerate}
For each task, we sequentially evaluate the full retargeting objective and the seven ablations (A1 to A7), and repeat the entire sequence three times. 
% The above tests are conducted twice with two human pilots.
For task 1 and 2, we use mean completion time to assess performance of pilots, as all trials are successful. 
For task 3, we use success rate for comparison. 
The manipulation results are summarized in Fig. \ref{fig:real_world_results}.
% Our real-world evaluation is a preliminary case study with only two human pilots. We plan to involve more pilots to capture variability among users for further validation.

\subsection{Analysis of Results}

We analyze the results of kinematic posture retargeting and real-world manipulations in \textit{four perspectives} corresponding to the designed objectives.
% : fingertip pinch, fingertip orientation, global hand pose, and joint position regularization.

% \textbf{Fingertip pinch}: A1 and A2 are the ablation of fingertip pinch term. A1, which removes the pinch term and excludes the relative position information between the thumb and primary fingers, results in significantly higher errors in both fingertip position and relative position to the thumb. It also fails in manipulation tasks involving pinch motions, as it cannot close the gap between thumb and index fingertips. 

% A2 shows minimal error in both metrics as it uses actual pinch distances without any rescaling. However, it achieves a low success rate in real-world manipulation tasks, primarily due to using actual pinch distance is vulnerable to finger tracking inaccuracy.

\textbf{Fingertip pinch}: A1 and A2 are the ablation of fingertip pinch term. 
The results show that:
1) removing the pinch term in A1 results in significantly higher errors in both fingertip position and relative position to the thumb (A1 in Fig. \ref{fig:leap_simulation_1}). This results from the potential conflict between the global fingertip position and the fingertip position relative to the wrist, due to different hand morphologies between the human and robot; 
2) in real-world manipulations, removing the pinch term in A1 leads to failure in tasks involving pinch motions, as it cannot close the gap between thumb and index fingertips (A1 in Task 3, Fig. \ref{fig:real_world_results}). This is also displayed in posture (a) in Fig. \ref{fig:posture_and_task};
3) using actual pinch distances without rescaling in A2 results in minimal error in both metrics (A2 in Fig. \ref{fig:leap_simulation_1}), but could lead to low success rate in real-world manipulation tasks, primarily due to that using actual pinch distance is vulnerable to human finger tracking inaccuracy (A2 in Task 3, Fig. \ref{fig:real_world_results}).

\textbf{Fingertip orientations}: A3 and A4 correspond to the ablation of fingertip orientation term. The results suggest: 1) DIP-to-tip vectors hold more explicit information of fingertip orientation than wrist-to-DIP vectors. Replacing the DIP-to-tip vectors with wrist-to-DIP vectors shows little advantage in fingertip orientation error over no orientation consideration (A4 in Fig. \ref{fig:leap_simulation_1}), 
as the DIP-to-tip vectors may be changed by fingertip relative position terms even though the wrist-to-DIP vectors are fixed;
2) in real-world manipulation Task 2, no consideration of fingertip orientations results in poor performance as the robot hand fails to replicate the finger bending motion of the human pilots to hook the mug handle (A3 in Task 2, Fig. \ref{fig:real_world_results}). In addition, it may occur that the finger bends in a wrong direction like posture (b) in Fig. \ref{fig:posture_and_task};
and 3) the missing of fingertip orientation information can also have negative impacts on precise manipulation tasks such as task 3, as inaccuracy in orientation leads to undesired contact normals with the object (A3 in Task 3, Fig. \ref{fig:real_world_results}).

\textbf{Global hand pose}: A5 and A6 are the ablation of global hand pose. The results show that:
1) determining the global hand pose by the thumb fingertip position term instead of wrist position term leads to a significantly lower error in fingertip global position (A5 in Fig. \ref{fig:leap_simulation_1}), primarily due to that using the thumb position term sacrifices in wrist position accuracy in exchange for better alignment of fingertips;
2) when using the exact wrist pose, removing the pinch term also leads to higher errors in fingertip relative position to the thumb  (A6 in Fig. \ref{fig:leap_simulation_1}), and removing the fingertip orientation term hampers finger bending (A6 in Task 2 in Fig. \ref{fig:real_world_results});
3) in real-world tasks, the choice of thumb fingertip position term and wrist position term does not bring up much difference (A5 in Fig. \ref{fig:real_world_results}), because human pilots can easily adjust the global hand position to eliminate the influence of fingertip global position inaccuracy. 

\textbf{Joint position regularization}: A7 and A8 removes the joint position regularization term. The corresponding results suggest that:
1) from the comparison between full and A7 in Fig. \ref{fig:leap_simulation_1}, joint position regularization term seems to have little impact in both simulation and real-world manipulation. However, without this term, it may occur that the hand displays undesired joint configurations like posture (c) in Fig. \ref{fig:posture_and_task}. The joint position regularization makes the retargeting result more truthful without negative impacts;
2) the comparison of A8 and A6 shows that when using wrist position term for global hand pose, the joint position regularization term adds to the errors in fingertip global position and relative position to the wrist and the thumb (A8 in Fig. \ref{fig:leap_simulation_1}). In contrast, when using thumb fingertip position, the joint position regularization term has little impact on position errors.

\subsection{Comparison with Existing Retargeting Approaches}

\begin{figure*}[tb]
    \centering
    \subfigure[]{
        \centering
        \includegraphics[width=\linewidth]{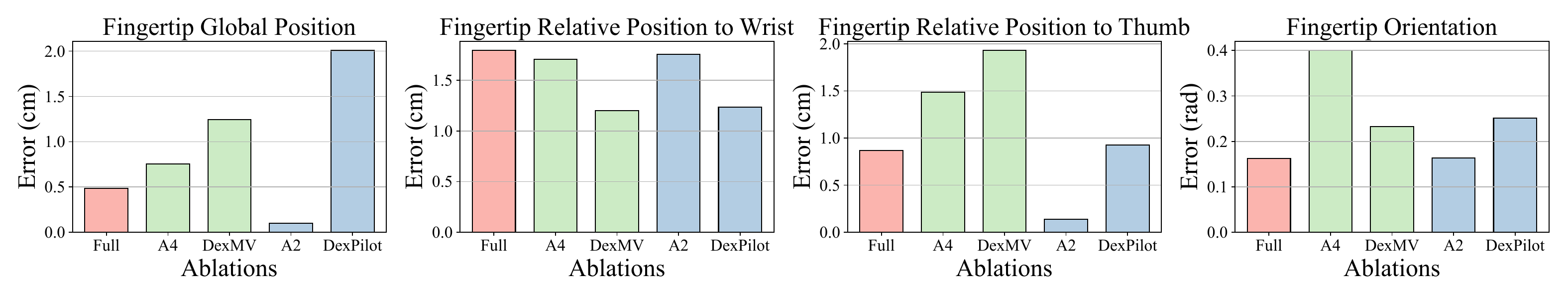}}
    \subfigure[]{
        \centering
        \includegraphics[width=0.6\linewidth]{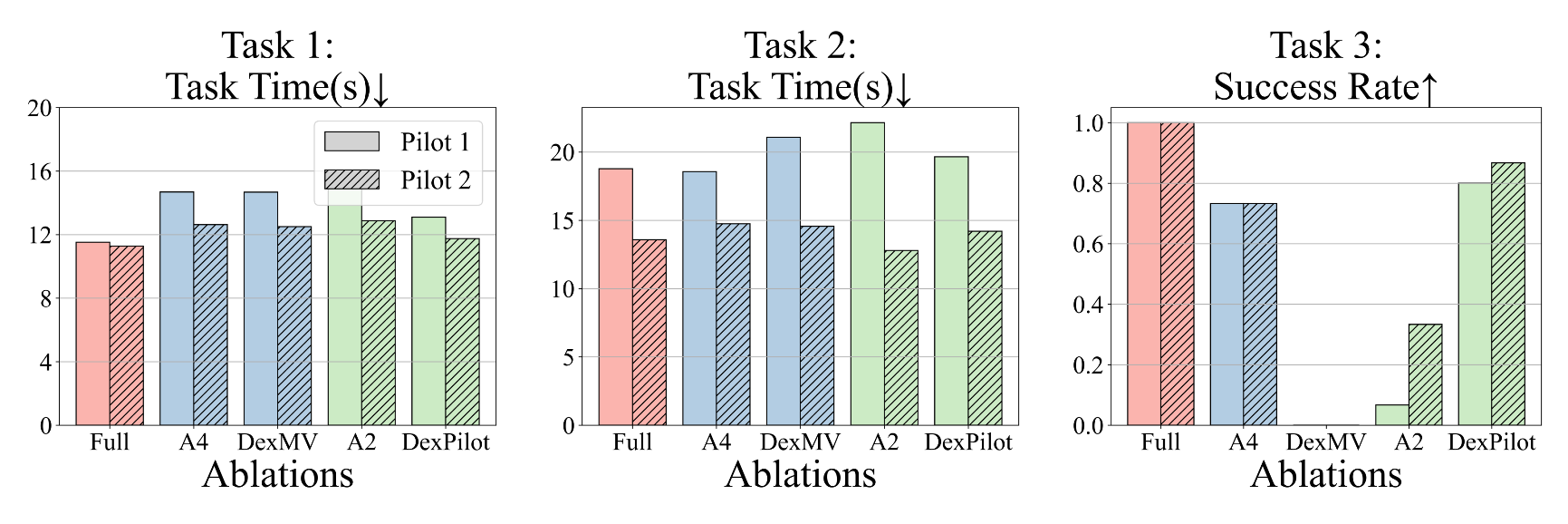}}
    \caption{(a) Results of kinematic posture retargeting on finger pinch trajectories. (b) Results of the real-world manipulations. Task 1 and 2 are assessed by task time, while Task 3 is evaluated by success rate. The bars represent the full objective, two ablations (A2, A4), DexMV, and DexPilot.}
    \label{fig:comparison_appendix}
\end{figure*}

In addition to the ablation studies, we also compare our full proposed method with two representative retargeting approaches, DexMV and DexPilot~\cite{qin2022dexmv,handa2020dexpilot}. 
In our ablation setting, A2 and A4 are designed to closely mimic the key design choices of DexPilot and DexMV respectively. 
Specifically, A2 uses the raw fingertip pinch distance in the pinch term;
unlike DexPilot, it does not include the fingertip distance rescaling function to ensure minimal distance between the thumb and a primary finger and force minimum separation distance between two primary fingers.
A4 replaces the vectors from DIP joints to fingertips with vectors from the wrist to DIP joints, which is similar to the fingertip orientation formulation in DexMV (we use DIP joints instead of the PIP joints used in the original DexMV).

In contrast, the pinch term in our full objective differs from DexPilot in two aspects. 
First, we use a continuous switching weight $s(d_i)$ based on a sigmoid function, instead of the discrete weights in DexPilot. 
Second, our distance rescaling function $l(d_i)$ varies continuously on $[\epsilon_2, \epsilon_1]$, while DexPilot directly clamps fingertip distances below a fixed threshold. 
As a result, in our objective, both the rescaled distance and its weight change smoothly as the human pinch distance crosses threshold $\epsilon_2$ and $\epsilon_1$, which leads to smoother and more stable motions during precise pinching in real-world teleoperation.

To further ensure a fair comparison beyond these approximate variants, we additionally implement DexMV and DexPilot by directly following their objective functions and hyper-parameter settings. 
For DexPilot, we follow the code in the dex-retargeting repository~\cite{qin2023anyteleop}.
Since their formulations are defined only for hand retargeting, we integrate them into our arm-hand retargeting pipeline by first applying their hand retargeting objective in the wrist frame and then separately solving the arm IK to track the resulting wrist pose.

We evaluate the full objective, the two ablations (A2 and A4), DexMV, and DexPilot on the LEAP-hand kinematic retargeting trajectories on finger pinch motions described in Sec.~III-A, and compute the same four metrics as in Fig.~\ref{fig:leap_simulation_1}: fingertip global position error, fingertip relative position to the wrist, fingertip relative position to the thumb, and fingertip orientation error. We also evaluate them on the same three real-world manipulation teleoperation tasks as Sec.~III-A.

Figure~\ref{fig:comparison_appendix} summarizes the quantitative kinematic posture retargeting and real-world manipulation results. The results show that, on the kinematic retargeting trajectories (Fig.~\ref{fig:comparison_appendix}(a)), the full objective consistently achieves lower fingertip global position error, fingertip relative position to the thumb, and fingertip orientation error compared to DexMV and DexPilot.

The real-world teleoperation experiments in Fig.~\ref{fig:comparison_appendix}(b) further validate our design choice. 
Across all three manipulation tasks, the full objective achieves the shortest or comparable task times and the highest success rates for both human pilots. 
DexMV and DexPilot, by contrast, generally lead to slower executions and more failures, especially on Task 3 which requires precise fingertip pinch. 
In particular, DexMV fails in every trial of Task 3 because it does not include the fingertip pinch term and therefore cannot close the gap between thumb and index fingertips, which is similar to the results of ablation A1 where we remove the pinch term.
These results demonstrate that the proposed objective not only improves kinematic retargeting quality but also leads to more efficient and reliable performance in real-world dexterous teleoperation compared to existing retargeting approaches.

Supplementary materials are listed in Table \ref{tab:supple}.

\begin{table*}[t]
\centering
\caption{Descriptions of Supplementary Materials}
\label{tab:supple}
\renewcommand{\arraystretch}{1.5}  % 设置行间距为默认的 1.5 倍
\scalebox{1.25}{
\begin{tabular}{ p{3cm} p{10cm} } 
\toprule
\textbf{Supplementary material} & \textbf{Description}  \\ 
\hline
Project website & 
\href{https://mingrui-yu.github.io/retargeting}{https://mingrui-yu.github.io/retargeting}
\\
% \hline
Appendix         &  
% Supplementary results and implementation details are provided in Appendix, which includes hyper-parameters of the retargeting algorithm, the detailed formulation of the pinch objective, and additional quantitative results of kinematic posture retargeting on another trajectory involving finger crossing motions and on the Shadow Hand.  The appendix is also available on the \href{https://mingrui-yu.github.io/retargeting}{project website}.
Supplementary results and implementation details are provided in Appendix, which includes additional quantitative results of kinematic posture retargeting on another trajectory involving finger crossing motions and on the Shadow Hand, explanation of the joint velocity regularization term, and implementation details such as hyper-parameter settings and discussion on runtime, latency, and robustness. The Appendix is also available on the \href{https://mingrui-yu.github.io/retargeting}{project website}.
\\
% \hline
Video             &   
The video of the real-world experiments is available on the \href{https://mingrui-yu.github.io/retargeting}{project website}, which demonstrates the real-world kinematic posture retargeting, three real-world manipulation tasks for evaluation, and additional trials on manipulation tasks of higher complexity.
\\
% \hline
Source code       &
The code is open-sourced on the \href{https://mingrui-yu.github.io/retargeting}{project website} (GitHub), which includes the implementation of the retargeting algorithm and the evaluation on kinematic postures in simulation. We provide a detailed guideline for setting up everything and to launch the evaluation. Users are encouraged to report issues via GitHub.            
\\
% \hline
Dataset           &   All human hand motion trajectories recorded by Apple Vision Pro in this study are provided on the \href{https://mingrui-yu.github.io/retargeting}{project website}. The format of the dataset is described by the instructions in the corresponding README file.                  
\\
% \hline
CAD files         &   The  CAD files of the fingertips and the URDF of the whole robot (i.e., Panda arm + Leap Hand + fingertips) are provided on the \href{https://mingrui-yu.github.io/retargeting}{project website}.                  
\\
\bottomrule
\end{tabular}
}
\end{table*}

\section{Conclusion}
This work analyzes the significance of different objectives in human-to-robot retargeting for dexterous manipulation through both kinematic posture retargeting and three representative real-world manipulation tasks.
The comprehensive results demonstrate that 
1) the fingertip pinch objective is crucial for manipulation tasks involving precise fingertip coordination; 
2) the fingertip orientation objective should be included for tasks sensitive to finger orientation rather than solely position; 
3) allowing wrist pose adjustment instead of using exact human wrist pose benefits the accuracy of fingertip poses; 
4) joint position regularization makes the retargeting postures appear more natural while having negligible negative impacts; 
and 5) all terms using the proposed formulation do not demonstrate conflicts and perform well in all tasks.
We also show in the experiments that the complete objective we purpose leads to more accurate and reliable performance than existing retargeting approaches.
We believe this study provides valuable insight for designing retargeting algorithms in future work on learning dexterous manipulation from humans or dexterous teleoperation.
In future work, we plan to extend the real-world evaluation to more teleoperators and tasks to better capture user variability and validate the generality of the proposed retargeting formulation.

% \clearpage

% \section*{Descriptions of Supplementary Materials}

% \clearpage

\bibliographystyle{IEEEtran}
\bibliography{IEEEabrv, ref}

% \clearpage

\clearpage

\appendix

\begin{figure*}[hb]
    \centering
    \subfigure[]{
        \centering
        \includegraphics[width=\linewidth]{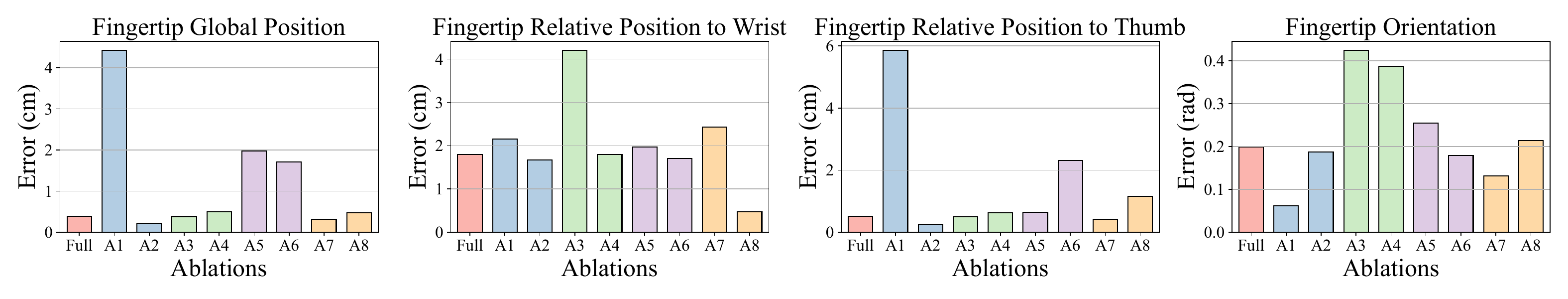}}
    \subfigure[]{
        \centering
        \includegraphics[width=\linewidth]{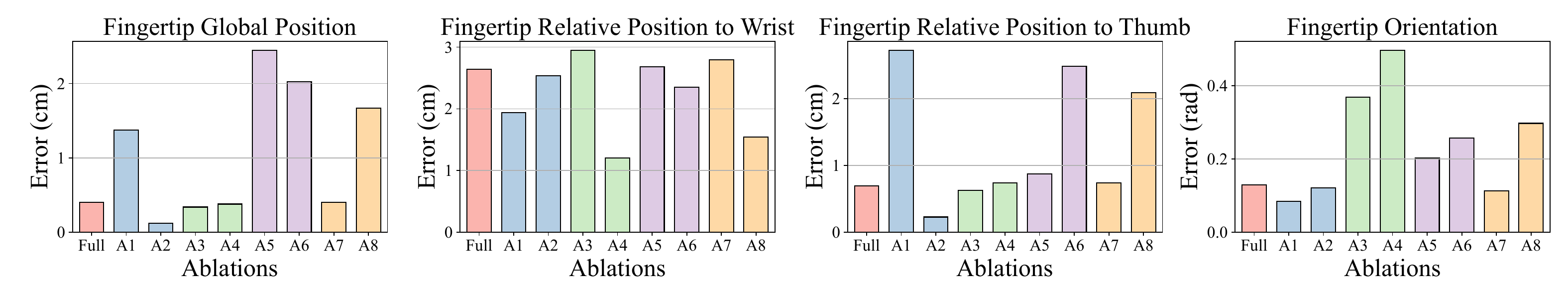}}
    \subfigure[]{
        \centering
        \includegraphics[width=\linewidth]{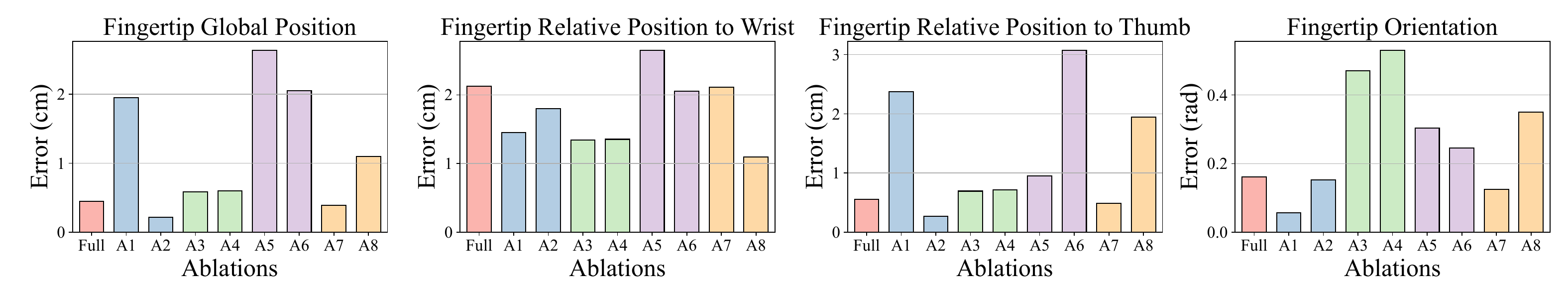}}
    \caption{Additional results on kinematic posture Retargeting. (a) Kinematic posture retargeting results using Leap hand on another trajectory involving finger crossing motion. (b) Kinematic posture retargeting results using Shadow hand on the same pinch motion trajectory as in the main text. (c) Kinematic posture retargeting results using Shadow hand on the trajectory involving finger crossing motion.}
    \label{fig:additional_kinematic_posture}
\end{figure*}

\section*{Additional Kinematic Posture Retargeting Results}
Three sets of additional quantitative results of kinematic posture retargeting on another trajectory involving finger crossing motions and on the Shadow Hand are shown in Fig. \ref{fig:additional_kinematic_posture}. Similar conclusions to the main text can be derived.

\section*{Explanation of Joint Velocity Regularization Term}

As introduced in Sec. II, the joint-velocity regularization term
\[
\mathcal{L}_{\text{vel}} = \sum_{j=1}^{m} w_{j}^{\rm vel} \left\| q_j - q_j^{\text{prev}} \right\|_2^2
\]
penalizes large changes in joint positions between consecutive timesteps and is intended to encourage smooth retargeted joint trajectories.

To illustrate the effect of this term, we compare the full objective in \eqref{eq:full_objective} with an ablation that removes $\mathcal{L}_{\text{vel}}$. 
We use a real-world teleoperation trajectory that involves finger-crossing motions with the index finger and generate the corresponding retargeted joint trajectories under both settings. 
The trajectories are evaluated on the three joints of the index finger (MCP, PIP, and DIP) on the LEAP Hand.

\begin{figure*}[htb]
  \centering
  \includegraphics[width=\textwidth]{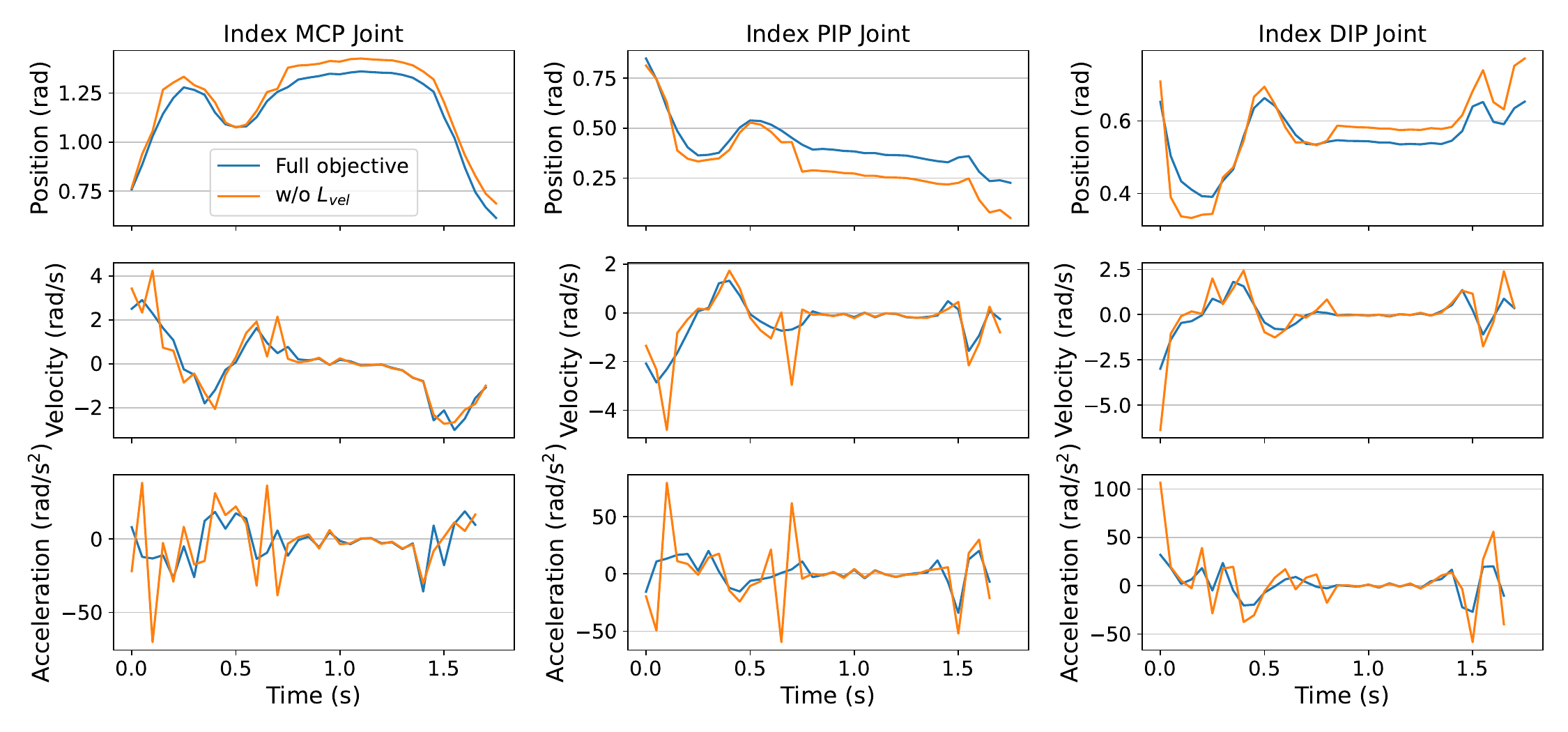}
  \caption{Index MCP/PIP/DIP joint position/velocity/acceleration profiles with and without $\mathcal{L}_{\text{vel}}$. For all three joints, the position trajectories are smooth and free of oscillations in both cases. However, removing $\mathcal{L}_{\text{vel}}$ consistently leads to larger peaks in the velocity and especially in the acceleration profiles, indicating sharper changes in joint motion during the finger-crossing motion.}
  \label{fig:joint_profile}
\end{figure*}

Figure~\ref{fig:joint_profile} shows the joint position, velocity, and acceleration profiles for the index MCP, PIP, and DIP joints, respectively, under the full objective (blue line) and the variant without $\mathcal{L}_{\text{vel}}$ (orange line). These results empirically support the importance of $\mathcal{L}_{\text{vel}}$ in suppressing high-frequency variations and reducing acceleration spikes, thereby improving the overall trajectory smoothness.

\section*{Implementation}
\subsection*{Hyper-parameter Setting}
The hyper-parameters used in the retargeting objective are listed in Table \ref{tab:hyper_parameters}.

\begin{table}[h]
\centering
% \vspace{-4mm}
\caption{Hyper-parameters}
% \vspace{-2mm}
\begin{tabular}{ll}
\toprule
\textbf{Hyper Parameter} & \textbf{Value} \\
\midrule
$\lambda_1$ & 10 \\
$\lambda_2$ & 0.1 \\
$\lambda_3$ & 1 \\
$\lambda_4$ & 10 \\
$\lambda_5$ & 10 \\
\midrule
\multirow{3}{*}{$w_j^{\text{pos}}$ (Leap hand)}
 & 0.5, $j=7,11,15,18$ \\
 & 0.1, $j=20$ \\
 & 0, else \\
 \midrule
 \multirow{3}{*}{$w_j^{\text{pos}}$ (Shadow Hand)}
 & 0.5, $j=9,13,17,22$ \\
 & 0.1, $j=26$ \\
 & 0, else \\
 \midrule
 \multirow{2}{*}{$w_j^{\text{vel}}$ (Leap hand)}
 & 0.1, $j=0 \sim 6$ \\
 & 0.01, $j=7 \sim 22$ \\
 \midrule
 \multirow{2}{*}{$w_j^{\text{vel}}$ (Shadow Hand)}
 & 0.1, $j=0 \sim 6$ \\
 & 0.01, $j=7 \sim 30$ \\
\bottomrule
\end{tabular}
\label{tab:hyper_parameters}
% \vspace{-5mm}
\end{table}

The weights $\lambda_1$–$\lambda_5$ reflect the relative importance of different retargeting objectives. We assign larger weights of 10 to the thumb-tip position, fingertip orientation, and pinch terms ($\lambda_1$, $\lambda_4$, $\lambda_5$), as these factors are crucial for dexterous manipulation and contact-rich pinching tasks. The weight of 0.1 for the wrist orientation term ($\lambda_2$) is relatively smaller as the objective aims to emphasize
fingertip tracking accuracy rather than wrist pose accuracy. The weight for the fingertip position term ($\lambda_3$) is set to 1 so that its contribution is of the same order of magnitude as the other terms when evaluated in typical retargeting scenarios.

For the weights of joint position and velocity regularization terms $w_j^{\text{joint}}$ and $w_j^{\text{vel}}$, index $j$ from 0 to 6 corresponds to the joints of the Panda arm, while indices $j = 7$ to $22$ and $j = 7$ to $30$ correspond to the joints of the Leap hand and the Shadow Hand respectively. Note that here we assume all DoFs of the Shadow Hand are actuated.

The joint position regularization weights $w_j^{\text{pos}}$ are determined according to the physical roles of different joints. 
For Leap hand, $j=7,11,15$ correspond to the abduction/adduction joints of index, middle and ring, $j=18$ corresponds to the DIP joint of the ring, and $j=20$ corresponds to the rotation of the thumb. 
For Shadow Hand, $j=9,13,17,22$ correspond to the finger movements of index, middle, ring and little finger in the palm plane, while $j=26$ corresponds to the rotation of the thumb.
Non-zero weights are assigned to these joints whose extreme values tend to produce unnatural or mechanically unfavorable configurations. 
For joints with non-zero position regularization, the pre-defined joint configurations are set to $\bar{q}_{j}=0$.

The joint velocity regularization weights $w_j^{\text{vel}}$ are chosen to penalize large changes in joint positions compared to previous timestep to encourage trajectory smoothness. In practice, we use larger $w_j^{\text{vel}}$ for the arm joints and slightly smaller values for the finger joints, as arm motions are more prone to sudden changes. As shown in Fig.~\ref{fig:joint_profile}, this choice effectively reduces acceleration spikes and leads to smoother joint trajectories.

In our implementation, we rescale the size of the human hand by a factor of 1.5 for the Leap hand and 1.0 for the Shadow Hand to address the size difference between human and robot hand.
In the real-world experiments, the retargeting control frequency is 20 Hz, and we use an exponential moving average with $\alpha_{\rm ema}=0.3$ to further smoothen the joint movements:
\begin{equation}
    \bm{q}_t = \alpha_{\rm ema}\cdot\bm{q}_t+(1-\alpha_{\rm ema})\cdot\bm{q}_{t-1}
\end{equation}

% \subsection*{\blue{Control Approach}}

% \blue{In the real-world experiments, we use joint-space position control for both the Panda arm and the robot hand. At each retargeting step (20 Hz), the retargeting optimizer outputs a desired joint configuration $\bm{q}_t$ for all actuated joints. This configuration is first upsampled to a 100 Hz command stream via interpolation and then sent to the hardware. 
% The LEAP Hand executes the control command via a built-in current-based PD controller.
% The Franka arm runs an impedance controller at a higher 1000 Hz with strict acceleration and jerk limits, so the 100 Hz joint targets are further interpolated to match the 1000 Hz control frequency before being executed by the arm controller.
% }

\subsection*{Discussion on Runtime, Latency, and Robustness}

\paragraph{Runtime and latency}

We profiled the computational cost of our retargeting pipeline on the LEAP-hand kinematic retargeting experiments. The full teleoperation loop runs at 20~Hz, corresponding to a frame time of 50~ms.
Within each frame, the nonlinear optimizer for the retargeting objective takes on average 33~ms, which is lower than the frame time and is suitable for real-time control.

Overall, latency in the system mainly comes from three stages: (i) hand pose estimation and streaming from the Vision Pro headset; (ii) retargeting optimization; (iii) communication latency between our teleoperation node and the robot; and (iv) the low-level control latency on the robot side. The Vision Pro perception module runs in a separate thread and continuously receives hand pose estimations, while our 20~Hz teleoperation loop always uses the latest available hand pose at the beginning of each control cycle. After that, the optimizer runs within about 33~ms and the resulting joint commands are immediately sent to the low-level joint controllers. 
The latency introduced by our retargeting module is below 50 ms, and the Vision Pro streaming introduces an additional latency of about 50 ms\cite{park2024avp}. Overall, the end-to-end latency of the system is approximately 0.15--0.25 s, which we empirically found sufficient for ensuring real-time teleoperation in all our real-world manipulation tasks.

\paragraph{Robustness to tracking noise and occlusions}

Accurate detection and tracking of human hand pose is key to dexterous teleoperation. We experimented with both a single RGB camera and the Vision Pro headset for hand pose detection. Compared to the RGB camera, Vision Pro provides a more flexible field of view and more accurate hand pose tracking, but it still inevitably exhibits some tracking noise, especially in precise motions such as pinching. In practice, we observed that when the human thumb and index fingertip are already in contact, the estimated fingertip distance from Vision Pro can remain slightly positive and exhibit small jitter over time. We specially design the pinch term in our objective to be more robust to tracking error to enable stable pinching: we rescale the estimated fingertip distances in a small range $[\epsilon_{2},\epsilon_{1}]$ to $[0,\epsilon_{1}]$ and clamp values below $\epsilon_{2}$ to zero (Sec.~III-A), so that sufficiently small estimated distances are treated as a closed pinch.

With respect to occlusions, Vision Pro can estimate a consistent hand pose under slight self-occlusion conditions, such as when a finger is partially occluded by the palm, as long as the operator keeps the hand within the headset’s field of view. This was sufficient for all three teleoperation tasks in our experiments, but for scenarios that require robustness under severe occlusions, Vision Pro may produce inaccurate hand pose estimation. Glove-based hand tracking systems will offer better robustness and be more occlusion-resistant in such cases\cite{wang2024dexcap, zhang2025doglove}.

\end{document}